\documentclass{article}
\usepackage{spconf,epsfig}

\usepackage{cite}
\usepackage[T1]{fontenc}
\usepackage[utf8]{inputenc}
\usepackage{fancyhdr}
\usepackage{ifthen}
\usepackage{caption}
\usepackage{float}
\usepackage{amsmath}
\usepackage{ wasysym }
\usepackage{ amssymb }
\usepackage{ upgreek }
\usepackage{algorithm}
\usepackage{algpseudocode}
\usepackage{pifont}
\usepackage{tikz}
\usetikzlibrary{arrows,matrix,positioning}
\usepackage{amsthm}
\usepackage{enumerate}
\usepackage{subfigure}
\usepackage{graphicx}
\usepackage{bbm}
\usepackage{wrapfig}
\usepackage{cancel}
\usepackage{hyperref}
\usepackage{pdfpages}
\usepackage{tabularx}
\usepackage{hyperref}
\usepackage{pgfplots}
\usepackage{color}
\usepackage{mathrsfs}
\usepackage{multirow}

\newtheorem*{theorem*}{Theorem}

\newcommand{\bpg}{\begin{paragraph}{}}
\newcommand{\epg}{\end{paragraph}}
\newcommand{\setpg}[1]{\begin{paragraph}{\bf Question #1:}}
\newcommand{\unsetpg}{\end{paragraph}\newpage}
\newcommand{\bit}{\begin{itemize}}
\newcommand{\eit}{\end{itemize}}
\newcommand{\beq}{\begin{equation}}
\newcommand{\eeq}{\end{equation}}
\newcommand{\beqn}{\begin{equation*}}
\newcommand{\eeqn}{\end{equation*}}
\newcommand{\beqa}{\begin{equation}\begin{aligned}}
\newcommand{\eeqa}{\end{aligned}\end{equation}}
\newcommand{\beqna}{\begin{equation*}\begin{aligned}}
\newcommand{\eeqna}{\end{aligned}\end{equation*}}

\newcommand{\enum}{\begin{enumerate}}
\newcommand{\enuma}{\begin{enumerate}[(a)]}
\newcommand{\eenum}{\end{enumerate}}
\newcommand{\norm}[1]{||#1||}

\DeclareMathOperator{\argmax}{argmax}

\newcommand{\lp}{\left(}
\newcommand{\rp}{\right)}

\newcommand{\matrixcolsep}[1]{\kern#1em\vline}

\newcommand{\boly}{\boldsymbol{y}}

\newcommand{\bolb}{\boldsymbol{\beta}}

\newcommand{\bol}[1]{\boldsymbol{#1}}

\newcommand{\twonorm}[1]{\norm{#1}_2}
\newcommand{\onenorm}[1]{\norm{#1}_1}

\newcommand{\makered}[1]{{\color{red}#1}}
\newcommand{\makegreen}[1]{{\color{green}#1}}
\newcommand{\makeblue}[1]{{\color{blue}#1}}

\definecolor{dkgreen}{rgb}{0,0.6,0}
\definecolor{gray}{rgb}{0.5,0.5,0.5}
\definecolor{mauve}{rgb}{0.58,0,0.82}

\algnewcommand\algorithmicinput{\textbf{Input}:}
\algnewcommand\INPUT{\item[\algorithmicinput]}
\algnewcommand\algorithmicoutput{\textbf{Output}:}
\algnewcommand\OUTPUT{\item[\algorithmicoutput]}

\usetikzlibrary{shapes,arrows,shadows}
\usepackage{amsmath,bm,times}

\makeatletter
\def\@seccntformatinl#1{\csname the#1dis\endcsname\hskip 1em\relax}
\makeatother

\begin{document}
%
\title{Localized Dictionary Design for Geometrically Robust Sonar ATR}

\twoauthors
{J. McKay, V. Monga\sthanks{Supported by Office of Naval Research, Arlington, VA, Grant 0401531}}{
Pennsylvania State University\\
Department of Electrical Engineering\\
University Park, PA}
{R. Raj}{
U.S. Naval Research Laboratory\\
Washington, DC}



\fancypagestyle{plain}{
\fancyhf[R]{}
\fancyfoot[L]{}
\fancyfoot[C]{}
\renewcommand{\headrulewidth}{0pt}
\renewcommand{\footrulewidth}{0pt}
}
\fancypagestyle{first}{
\fancyhead[R]{}
\fancyfoot{}
}

\pagestyle{plain}

\maketitle
\thispagestyle{first}
\begin{abstract}
Advancements in Sonar image capture have opened the door to powerful classification schemes for automatic target recognition (ATR).  Recent work has particularly seen the application of sparse reconstruction-based classification (SRC) to sonar ATR, which provides compelling accuracy rates even in the presence of noise and blur.  However, existing sparsity based sonar ATR techniques assume that the test images exhibit geometric pose that is consistent with respect to the training set.  This work addresses the outstanding open challenge of handling inconsistently posed Sonar images relative to training.  We develop a new localized block-based dictionary design that can enable geometric robustness. Further, a dictionary learning method is incorporated to increase performance and efficiency.  The proposed SRC with Localized Pose Management (LPM), is shown to outperform the state of the art SIFT feature and SVM approach, due to its power to discern background clutter in Sonar images.
\end{abstract}

\section{Introduction}
\begin{figure*}[t]\centering
\includegraphics[width=.59\textwidth]{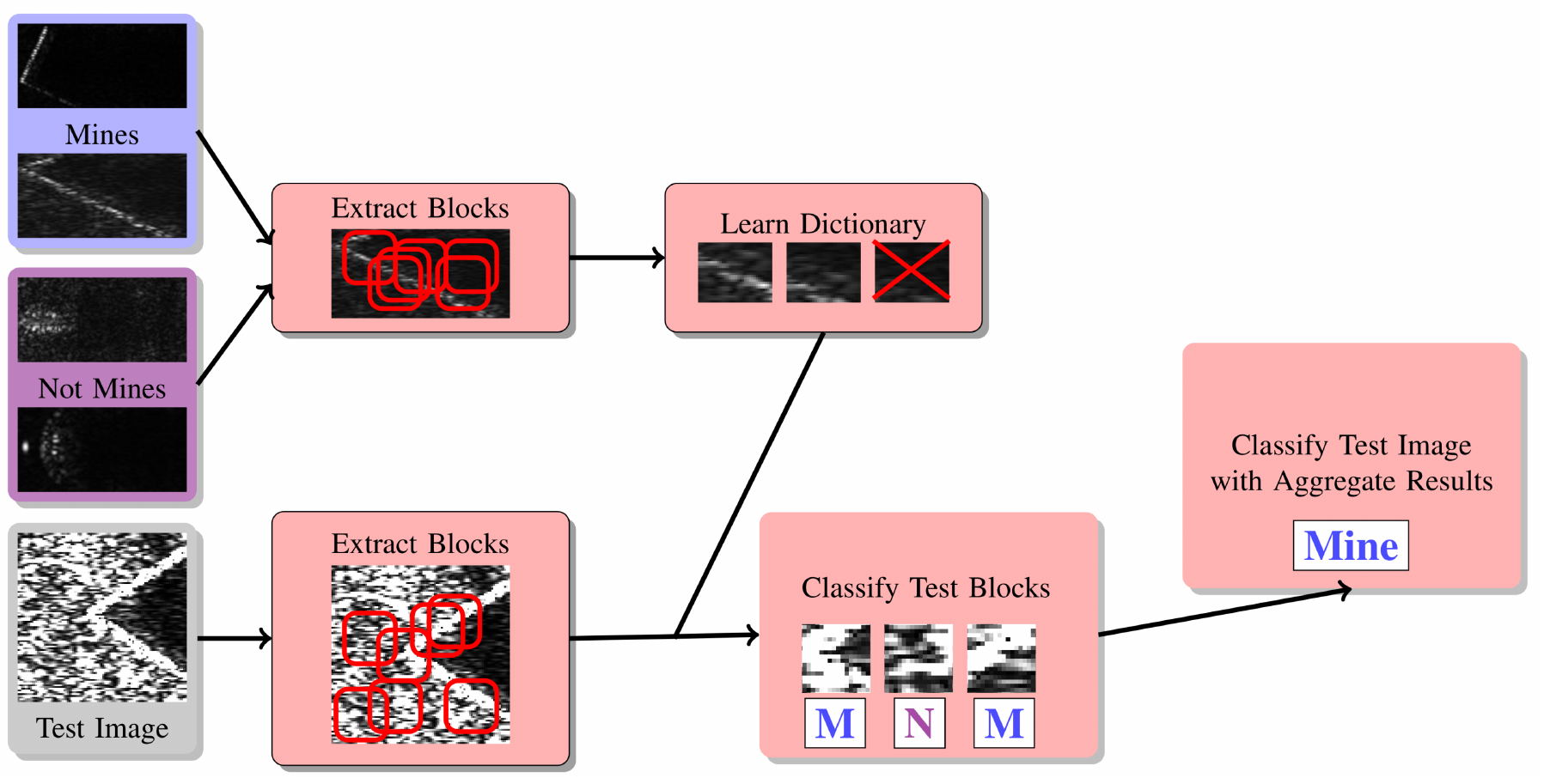}
\caption{SRC with localized pose management (LPM) for Sonar ATR.}\label{schema}
\end{figure*}

The threat of mines and other harmful underwater devices have made the problem of object identification via automated underwater vehicles (AUVs) a vital area of study for both military and commercial parties.  These machines, which offer greater mobility and safety over a human piloted submersible, can be used to detect targets of interest \cite{stack2011automation}.  Two ways for AUVs to do this are active target recognition where a human assists in the classification and automated target recognition (ATR).  While the former has its benefits, there are times where AUVs cannot incorporate human interactions into its classification.  For this reason, we look to investigate Sonar ATR.

Recent work in Sonar ATR has demonstrated the potential for sparse reconstruction-based classification (SRC) in this field \cite{kumar2012object} \cite{mckay2015discriminative} \cite{fandos2009sparse}.  SRC has been widely popular in computer vision circles since its introduction with \cite{wright2009robust} in 2009 because of its ability to perform even when pressed with occlusion and noise.  The authors of \cite{mckay2015discriminative} show how this characteristic makes SRC particularly attractive to Sonar ATR given how common noise is in this setting.  In addition to this, they show that SRC can thrive even when under significantly constrained training settings making it even more fitting given how large the training sample has to be in Sonar ATR to incorporate the different looks objects have at different angles.

One issue with SRC methods are their ability to handle images whose targets are not aligned in the same manner as the training images.  In the manner \cite{wright2009robust} and \cite{mckay2015discriminative} execute their SRC scheme, targets need to have the same location and dimension, reducing flexibility in real-world scenarios.  We look to address this issue of inconsistently-posed test images via localized pose management (LPM)
which exploits localized geometric information present in the image by sampling the global image with multiple sub windows and uses these to initialize the input to a well-known dictionary learning algorithm. It has been empirically established in diverse application domains \cite{srinivas2013shirc} \cite{lee1999nature} that localized features (such as corners and edges) reveal more discriminatory information than global geometric counterparts.  Additionally, geometric manipulations, such as transformations between geometric structures, are easier to handle at the local level.

In the following section, we give a summary of what SRC methods are and how they are implemented.  Next, we describe LPM in detail with focus on its dictionary learning step.  Lastly, we present the results of several experiments comparing our SRC with LPM method against a popular SIFT feature SVM using a dataset of Sonar images provided by the Naval Surface Warfare Center.

\section{SRC}\label{SRC}

Building on the success of sparsity-based methods in compressive sensing \cite{candes2008introduction}, \cite{wright2009robust} present SRC, a linear modeling framework that offers essentially no formal training and robust classification rates even when pressed with noise and blurring.  Their work starts by constructing a class-specific dictionary, $D$, using the available training images, i.e.
\beqna
D    =\begin{pmatrix}
\makered{D_1} & \makegreen{D_2} & \cdots & \makeblue{D_K}\\
\end{pmatrix}
\eeqna
where $D_j$ represents the dictionary of vectorized training images corresponding to class $j$.  With $D$ we can classify a vectorized test image $\boly$ by solving for $\bolb$
\beqa\label{eq:relax}
&\min\onenorm{\bolb}\\
&\text{  s.t. }\twonorm{\boly -D\bolb}<\varepsilon
\eeqa
where the $\ell_1$ typically induces the sparsest solution for $\bolb$ \cite{donoho2006most}.  There are several options with which to solve the above problem and the review by \cite{yang2010fast} presents a comprehensive overview.  We found a L1LS method to produce the most satisfactory results for our work.

\cite{mckay2015discriminative} showed that it is possible produce compelling classification rates on consistently-posed test images of Sonar image using SRC.  That said, in real-world cases the ability to collect targets all arranged in geometrically ideal positions is difficult if not infeasible.  For this reason, we present the following algorithm to handle pose diversity for SRC in Sonar ATR.

\section{SRC with Localized Pose Management}

To use SRC with geometrically diverse Sonar ATR settings, we develop a localized block-based approach.  For our method, images are segmented into several $M$ by $N$ blocks which are then used as the training images for the dictionary $D$ of \eqref{eq:relax}.  Each of the small blocks are assigned the label of whichever class the original larger image has.  The test image $\boly$ is then also segmented into blocks and each one is tested against $D$.  We use the term ``block'' is used instead of ``patch'' to highlight the fact that we have no intention to apply any feature transformation to the sub-images.

There are several routes to determining the class of the test image once all of its blocks have individually been classified via SRC.  A majority vote approach where the test image is assigned the most common class amongst its blocks is one of the simplest, but is also highly susceptible to misclassification in cluttered images with prominent background features.  Another method that has proven to yield the best results in our own experiments is a tailored maximal likelihood approach \cite{kittler1998combining}.  To understand it, consider the following:  let $\boly_{i}$ be an extracted test block and $\bolb_{i}$ be its coefficient vector found via SRC corresponding to the dictionary $D$.   Define $r_{i,k}$ as
\beqa\label{eq:res}
r_{i,k}=\frac{1}{\twonorm{\boly_i - D\bol{\delta}_k(\bolb_i)}}
\eeqa
where $\bol{\delta}_k(\bolb_i)$ is a vector that holds all the values of $\bolb_i$ corresponding to class $k$ and presents zero for all other entries.  The probability that $\boly_i$ belongs to any of the $K$ classes is
\beqa
p_i^k = \frac{r_{i,k}}{R_i}\text{ where }R_i=\sum\limits_{k=1}^K r_{i,k}
\eeqa
Thus, the maximal likelihood estimate of the class of $\boly$ is
\beqa\label{eq:maximal}
\argmax\limits_{k=1,\dots,K} \ln\lp \prod_{i=1}^I p_i^k\rp
\eeqa
where $I$ is the number of blocks extracted from the test image.

\begin{figure}[b]\centering
\includegraphics[width=4cm]{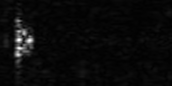}
\includegraphics[width=4cm]{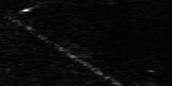}
\caption{Sonar images of two differently oriented cylinders.}\label{difangles}
\end{figure}

This strategy as a whole provides a straightforward way to use SRC with test images whose target is not aligned with the training and/or has dimension different from the training.  In the context of Sonar ATR, this approach offers a translationally invariant method by which to use SRC without any rotationally invariant confusion.  We note this as Sonar image capture can render quite different images for the same shaped object depending on the angle of the object to the device collecting the data, as figure \ref{difangles} shows.  Therefore, SRC with LPM is structured in such a manner to adhere to the constraints that Sonar imaging imposes.

An outcome of a dictionary design that concatenates many local block images is the issue of handling the very large matrix $D$.  The computational stress of SRC using every $M$ by $N$ training block can cause the process to be untenable for most machines, making any approach that reduces the size of $D$ valuable.  Additionally, it is within reason to believe that there is redundancy within each class' blocks.  In \cite{mckay2015discriminative}, the authors found that SRC can work in Sonar even when the dictionary lacks such redundancy so these essentially repeated blocks are unnecessary.  For these reasons, we look at a dictionary learning procedure as a justifiable strategy to condense $D$.

There are several different dictionary learning methods for which we could consider.  The implementation that \cite{vu2015histopathological} outlines has proven to yield highly robust dictionaries with relatively modest computational stress.  Their approach entails selecting a random assortment of training samples that are then fed into the Online Dictionary Learning (ODL) algorithm from \cite{mairal2009online} to be further refined.  ODL specializes in minimizing dictionaries to a condensed, discriminative form intended for sparsity-based applications.  The whole process serves as a structured means to overcoming our dictionary redundancies problems and, as we will see in section \ref{exper}, can perform well in seeking out mines in Sonar images.

Figure \ref{schema} provides a diagram of how the LPM strategy follows from the block extraction to the final classification of a test image.

\section{Experiments}\label{exper}

To test SRC with LPM with Sonar ATR, we used a dataset provided from the Naval Surface Warfare Center of authentic synthetic aperture Sonar (SAS) captures of 13 backgrounds with 4 separate shapes simulated in various arrangement.  Based on similarities, we divided the data into two categories:  mine-like and non-mine-like.  We used 40 inconsistently-posed test images, twenty per class, for our experiments.

 First, we looked to show how much the dictionary learning step of our SRC with LPM impacts classification.  We did so by performing a similar block-based scheme that randomly chose every element of its dictionary without filtering through a dictionary learning step.  Additionally, we show how the SRC method alone performs on our test images to give a baseline understanding of why a procedure to handle geometric diversity is needed.

All the SRC methods used 18 training images.  The two block-based approaches used 18 samples of 60 by 20 pixel blocks from each training image and involved the maximal likelihood scheme of \eqref{eq:maximal}.  This was implemented using the results 30 test blocks extracted from the test images.

\begin{figure}[t]\centering
\includegraphics[width=.16\textwidth]{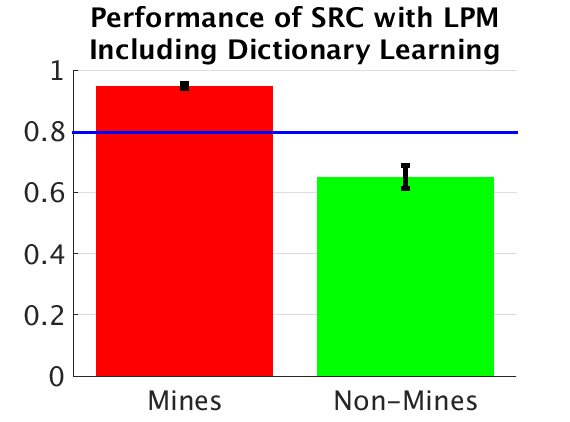}\hspace{-.3cm}
\includegraphics[width=.16\textwidth]{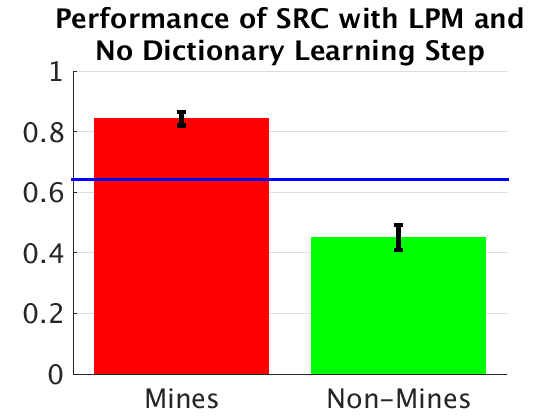}\hspace{-.3cm}
\includegraphics[width=.16\textwidth]{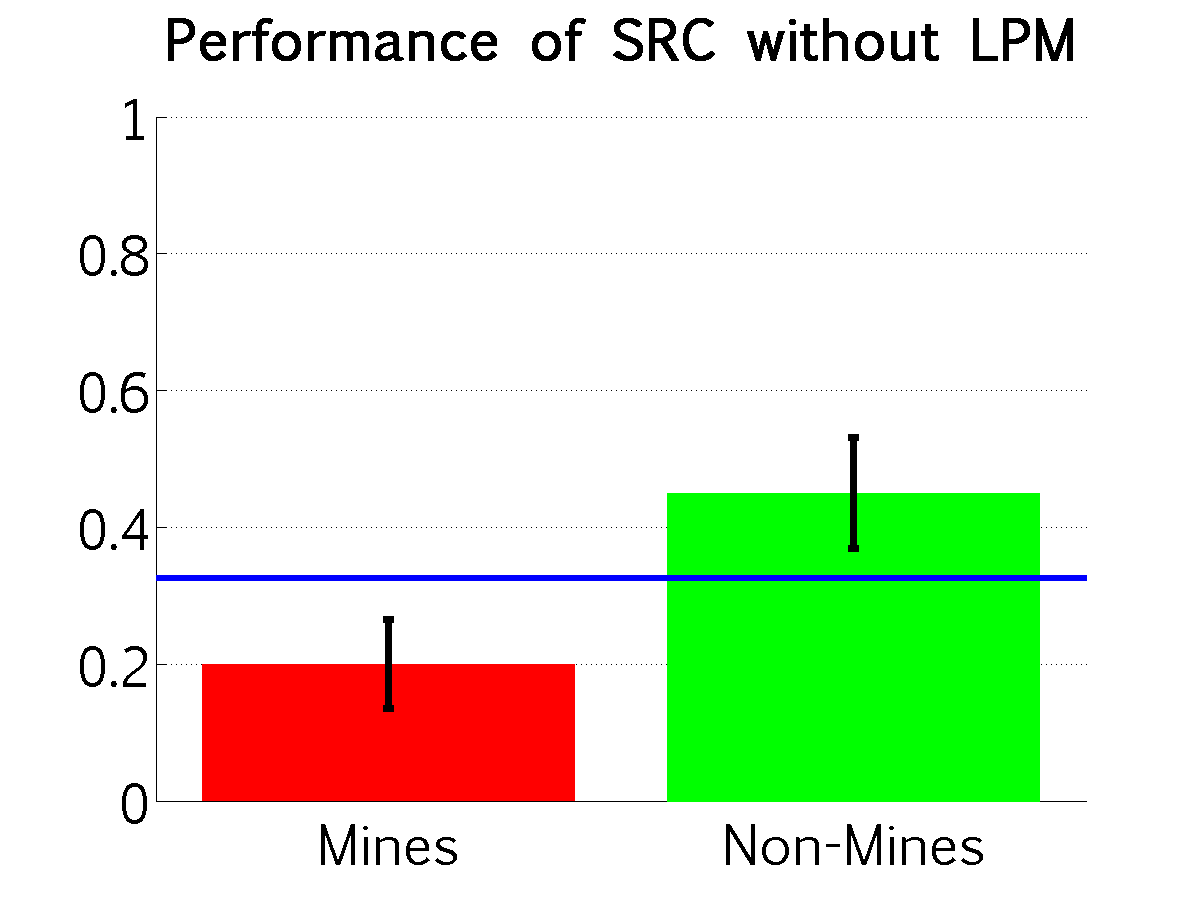}
\caption{SRC with LPM (left), LPM without dictionary learning (center), and without LPM (right); standard deviations shown.}\label{dict}
\end{figure}

As figure \ref{dict} depicts, the dictionary learning step provides a 15\% increase (79\% classification rate vs. 64\%) in accuracy over random sampling alone.  The benefits are not class-specific either as both mines and non-mines saw jumps in performance with dictionary learning.  This non-trivial result demonstrates how powerful ODL can increase the viability of SRC with LPM.

This said, even the randomized sampling approach performed better than the straight forward application of SRC on mis-aligned targets.  SRC alone performed poorly, yielding a 33\% accuracy rate.  Given how SRC works, it makes sense that it would do so poorly.  Without it or a similar method, SRC is poorly equipped to tackle real-world classification problems alone.

Next, we present how well SRC with LPM performs when compared to a popular image classification technique, SIFT feature SVM.  In \cite{zhu2014model} and \cite{mckay2015discriminative} the authors implemented SIFT feature SVMs towards Sonar ATR, the former of which in to handle pose-diversity.  For this reason, we used this algorithm on our test images to provide context for SRC with LPM.  Experiments involved 25 training images for the SRC with LPM and 50 for the SIFT feature SVM.  Our approach used 15 samples of 60 by 20 pixel blocks for the dictionary and 30 blocks taken from each test image for classifications.

\begin{figure}[b]\centering
\includegraphics[width=.22\textwidth]{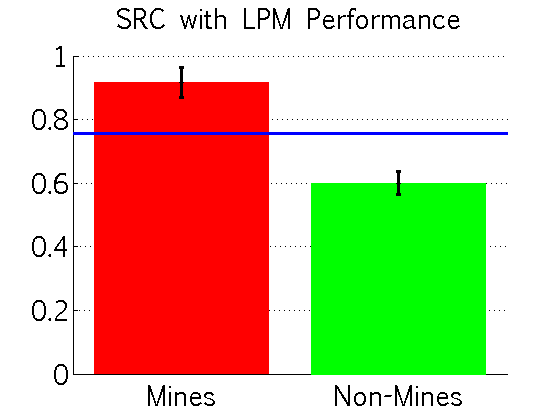}
\includegraphics[width=.22\textwidth]{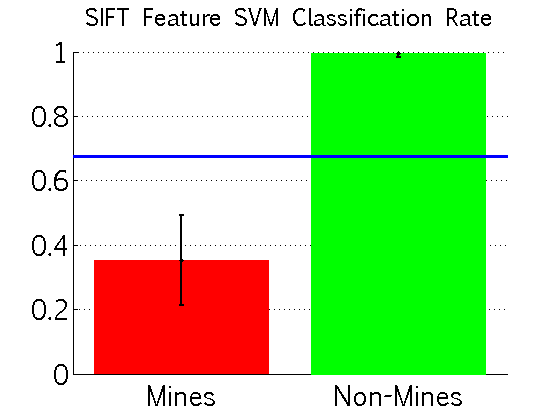}
\caption{SRC with LPM (left) vs. SIFT Feature SVM (right).  SRC used 25 training samples and the SVM used 50.  Standard deviation of trials shown.}\label{srcvssift}
\end{figure}

 Figure \ref{srcvssift} shows that our SRC with LPM outperforms the SIFT SVM method in overall classification rate (79\% vs. 68\%) and correctly identifying mines (90\% to 38\%).  It appears as though the SIFT SVM has a hard time discerning background clutter from the rounded edges of the non-mines, making for a high rate of false negatives.  For Sonar ATR, this tendency to miss on potentially threatening objects could be disastrous.  On the other hand, the SRC with LPM was able to present high mine-object classification rates with half the training of the SIFT SVM.  This further confirms the work of \cite{mckay2015discriminative} in showing how SRC can thrive even in limited training.

Lastly, we considered images with noise.  The process of Sonar image capture, especially SAS, can be susceptible to fair amounts of noise \cite{hayes2009synthetic}.  Thus, Sonar ATR methods have to show a certain degree of resiliency to this hindrance in order to prove its merit in real-world settings.  For the following, we added varying intensities of salt and pepper noise to the test images and saw how each method performed.

\begin{figure}[t]\centering
\includegraphics[width=.4\textwidth]{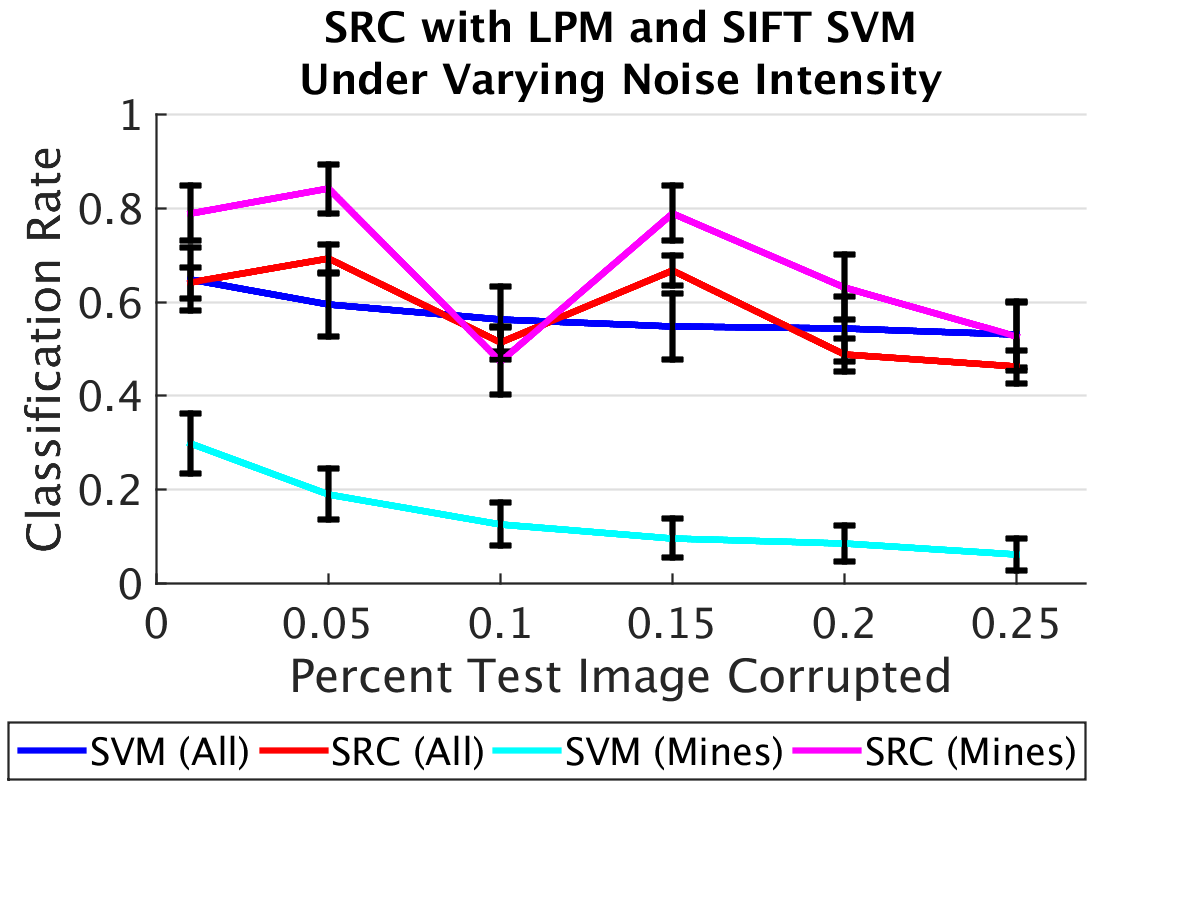}
\caption{SRC with LPM vs. SIFT feature SVM with noise.}\label{noised}
\end{figure}

\ref{noised} shows is that the SIFT feature SVM suffers great difficulty in classifying the mine-like objects while the SRC with LPM is able to still retain classification rates above 50\% for the same targets, even under 25\% pixel corruption.  The overall rate for the SIFT feature SVM is buoyed by its non-mine classification, but its trouble with mines is highly problematic.  The SRC with LPM seems some impact given noise but its resiliency in avoiding substantial false negatives gives it a great deal of value in real-world settings.

\bibliographystyle{IEEEbib}
\bibliography{refOceans2015New}

\end{document}